\pdfoutput=1

\documentclass[11pt]{article}

\usepackage{acl}

\usepackage{times}
\usepackage{latexsym}
\usepackage{graphicx}
\usepackage{booktabs}
\usepackage{multirow}
\usepackage{makecell}

\usepackage[T1]{fontenc}

\usepackage[utf8]{inputenc}

\usepackage{microtype}

\newcommand\blfootnote[1]{%
  \begingroup
  \renewcommand\thefootnote{}\footnote{#1}%
  \addtocounter{footnote}{-1}%
  \endgroup
}

\title{ 
Design Challenges for a Multi-Perspective Search Engine}

\author{Sihao Chen* \hspace{1cm} Siyi Liu* \hspace{1cm} Xander Uyttendaele \\ \bf{ Yi Zhang \hspace{1cm} William Bruno \hspace{1cm} Dan Roth} \\
University of Pennsylvania \\
\small \texttt{\{sihaoc, siyiliu, xanderu, yizhang5, wwbruno, danroth\}@cis.upenn.edu}}

\begin{document}
\maketitle

\newcommand{\dr}[1]{\textcolor{red}{[DR: #1]}}
\newcommand{\drc}[1]{\textcolor{red}{#1}}
\newcommand{\sihao}[1]{\textcolor{purple}{[Sihao: #1]}}
\newcommand{\siyi}[1]{\textcolor{blue}{[Siyi: #1]}}

\begin{abstract}

Many users turn to document retrieval systems (e.g. search engines) to seek answers to controversial or open-ended questions. However, classical document retrieval systems fall short at delivering users a set of \emph{direct} and \emph{diverse} responses in such cases, which requires identifying responses within web documents in the context of the query, and aggregating the responses based on their different \emph{perspectives}. 

The goal of this work is to survey and study the user information needs for building a \emph{multi-perspective search engine} of such. We examine the challenges of synthesizing such language understanding objectives with document retrieval, and study a new \emph{multi-perspective} document retrieval paradigm. We discuss and assess the inherent natural language understanding challenges one needs to address in order to achieve the goal. Following the design challenges and principles, we propose and evaluate a practical prototype pipeline system.
We use the prototype system to conduct a user survey in order to assess the utility of our paradigm, as well as understanding the user information needs when issuing controversial and open-ended queries to a search engine.
\end{abstract}

\section{Introduction}


    \blfootnote{* Equal Contributions}
    In the past two decades, web search has developed as a ubiquitous way for users to retrieve web information. Typically, a web search query is driven by a specific intent, or information need, i.e. the users' desire to seek answers to an open-ended question \cite{shneiderman1997clarifying, rose2004understanding}. Among the various types of questions, one of the most challenging types, from the perspective of designing an information retrieval (IR) system, is \emph{debate-worthy}, \emph{controversial questions}, for which an objectively-true answer or consensus seems impractical to reach. Rather, explicit or implied responses with \emph{different perspectives}, can be found within topically-related web documents.
    
    \begin{figure}[t]
    \centering
    \includegraphics[width=\linewidth]{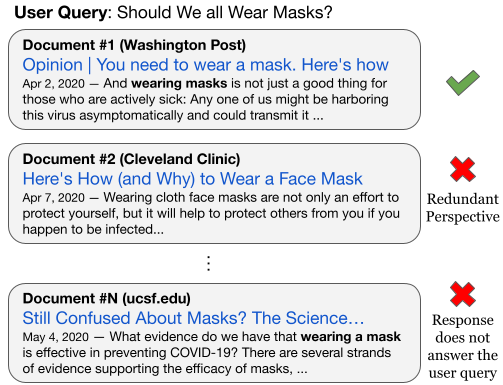}
    \caption{Example retrieved articles with snippets from a search engine with respect to the query ``Should we all wear masks'' dated from April to May 2020, during which mask wearing was a more controversial topic. The retrieved articles contain redundant information, or often the extracted summary \emph{response} does not answer the user query directly. }
    \label{fig:google_search}
    \end{figure}

    Classical IR systems fall short at delivering responses with a diverse set of perspectives to the user query. Instead, they typically provide a ranked list of references to relevant but not necessarily trustworthy web documents, with limited justifications on what part(s) of the document could answer the user query \cite{metzler2021rethinking}. We argue that a document retrieval system should ideally be equipped with the ability to extract or generate such responses from documents conditioned on the user query. Furthermore, a retrieval system should recognize the semantic difference of responses in cross-document settings, and in turn organize and deliver a set of documents from diverse perspectives. These objectives can inherently be formulated into a suite of natural language understanding tasks \cite{chen-etal-2019-seeing, lamm2020qed}.

    The goal of this paper is to discuss and validate the users' information need of \emph{direct and diverse} responses when issuing controversial, debate-worthy queries to a search engine. We examine the ideas of formulating such objectives as natural language understanding tasks, and incorporate them in a novel document retrieval paradigm. Through the lens of key design principles of a task-specific search engine \cite{rieger2009search}, we first discuss the challenges in designing such a multi-perspective search engine.  We introduce a paradigm that revolves around the abstraction of \emph{responses}, i.e. a text segment that serves as direct answer to users, and \emph{perspectives}, the argument(s) expressed in a response in the context of a given user query \cite{chen-etal-2019-seeing}. 
    
    To assess the utility of the proposed paradigm, we develop and demonstrate a prototype open-domain news search engine. Given a user query of controversial topics, the search engine retrieves news articles, identifies/generates responses within the articles along with evidence paragraphs, and categorizes the responses according to their perspectives differences. With the prototype, we conduct a user study in comparison to a general-purpose search engine. With the goal of understanding the user needs related to controversial queries, we discuss and analyze the advantages/disadvantages of our prototype through the user study responses. We then summarize and offer insights on the technical challenges and trade-off to cater to the specific user information needs with respect to controversial queries. We hope our findings will facilitate future research on this topic. 
    
    In summary, our contributions are as follows:
    \begin{enumerate}
        \item We study a document retrieval paradigm with the objective of delivering \emph{direct} and diverse summary responses to user query. We outline a list of natural language understanding challenges to achieve the goal.
        \item We demonstrate and evaluate a prototype multi-perspective search engine \footnote{Demo: \url{www.multiperspectivesearch.com}} \footnote{Code is released at \url{www.github.com/CogComp/multi-persp-search-engine}} that retrieve, organize and summarize the perspective behind news articles. 
        \item We conduct a user study with our prototype against a popular search engine to assess the utility of our paradigm, as well as to understand the information needs by users with respect to controversial, debate-worthy queries. 
    \end{enumerate}
    
\section{Related Work}
The importance of retrieval result diversity has long been recognized by the information retrieval community \cite{goffman1964searching, clarke2008novelty}. The problem has mostly been studied in the context of topic or entity ambiguity in the user query \cite{agrawal2009diversifying}, and most solutions focus on resolving the ambiguity and uncovering the user intent, or hierarchy of the target topics \cite{sakai2011evaluating, dang2013term, hu2015search, wang2017search}. The most notable hypothesis for this line of work, which our study also investigates, is the importance of modeling the inter-dependencies among documents during retrieval. The TREC 2009 Web track's diversity task \cite{clarkeoverview} introduces an evaluation protocol where the relevance of next retrieval item is dependent on the previous result. 

Our study takes a different angle from query ambiguity, and hypothesize that in the case of debate-worthy query, users wish to see \emph{direct} responses from diversified perspectives in the retrieved documents \cite{metzler2021rethinking}. Such objectives require recognizing the argumentation structure of a document \cite{stab2014identifying, ein2020corpus}, and more importantly comparing the semantics and implications of the arguments made cross documents \cite{chen-etal-2019-seeing, bar2020arguments}. 
\citet{chen2019perspectroscope} formulate a argument retrieval problem that conceptually resembles our formulation. In their task, a system is expected to return single-sentence arguments from different perspectives given a user query. Our study instead focuses on the more practical and challenging setting of document-level retrieval. 

The task of question answering (QA) aims to identify \emph{direct} answers to users questions expressed in natural languages from either close- or open-domain information resources \cite{kolomiyets2011survey}. Naturally, document retrieval has been an essential step during answer candidate generation \cite{chen2017reading}. Among various types of QA tasks, Machine reading comprehension \cite{rajpurkar2016squad, khashabi2018looking} and open-domain QA \cite{yang-etal-2015-wikiqa,joshi2017triviaqa,dunn2017searchqa} resemble our problem closely, as both involves identifying a concise response from large- to web-scale corpora. Under such settings, researchers have found benefit in jointly learning QA and retrieval tasks \cite{das2018multi}. The key difference between most QA tasks and our problem is that objective correctness does not exist for most controversial queries, and so they cannot be responded with a single correct answer. Instead, there exist dimensions such as stance \cite{bar-haim-etal-2017-stance} and persuasiveness \cite{carlile2018give} to measure the semantic difference between relevant and arguably equally valid arguments.  To this end, recent researchers have attempted to extract or generate explanations alongside search results \cite{lamm2020qed}.

\begin{figure*}[t]
    \centering
    \includegraphics[width=\linewidth]{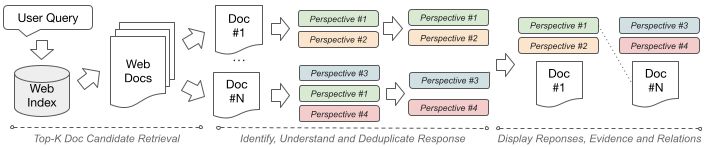}
    \caption{An conceptual illustration of our multi-perspective document retrieval paradigm, as discussed in $\S$~\ref{ssec:pipeline}. Instead of document, our retrieval pipeline revolves around \emph{direct responses} within articles. The responses are extracted or generated from individual candidate documents, and then they are re-ranked, de-duplicated and categorized based on the implied perspectives (e.g. different colors in the figure) across documents. Finally, the responses from \emph{diverse perspectives}, supporting evidence in context and relation between results are visualized in a parallel fashion.}
    \label{fig:pipeline}
\end{figure*}

\section{Design Challenges}
\label{sec:design_challenge}
We start by discussing the needs of users when issuing controversial and debate-worthy queries to a search engine. \citet{rieger2009search} introduces three key factors in building a task-specific search engine that caters to user's information needs. We introduce each factor below, and discuss the connections and implications to the case of controversial, debate-worthy user queries.

\textit{Satisfaction} refers to overall success of the retrieved results in providing help to the user's specific search need. Here there are two levels of user satisfaction to consider. The first dimension is whether the retrieved results \emph{contain} the answer that user needs. The second more challenging yet crucial dimension is whether search engine is able to highlight and display the answer \cite{metzler2021rethinking}. In the context of controversial queries, identifying direct answers from retrieved documents seems essential, as it will greatly reduce the amount of readings user need to do before deriving the answer. However, automating the process comes with technical challenges, and such automation often ties closely with concerns of model biases, as we will further discuss in $\S$~\ref{sec:discussion}.

\textit{Search Intent} refers to the category of webpages or answers that a user is interested to see when issuing a query. In the case of controversial queries, as users typically do not know a priori the potential aspects or perspectives that could lead to the answer in their context, organizing and highlighting a \emph{comprehensive} set of results with \emph{diverse} perspectives in a parallel fashion becomes important in this case \cite{chen2019perspectroscope}. 

\textit{Trust} is originally defined by \citet{pan2007google} as users' trust in a search engine's capability of retrieving relevant results to the query. However, as concerns over the veracity of web information grow \cite{thorne2018fever}, user trust revolves more closely around the trustworthiness of the sources and content \cite{pasternack2013latent}. For controversial queries, as multiple perspectives exist for every issue, whether each perspective is corroborated by evidence from trustworthy sources becomes the key to ensure user trust \cite{chen-etal-2019-seeing}. 

In the following sections, we take a closer look into the specific problems we discussed for the three factors, and the challenges in formulating NLP tasks to address the problems.  

\subsection{Identifying Direct Responses to Query}
\label{ssec:identify_response}
From the classical view of information retrieval, identifying responses, or \emph{explanations} \cite{lamm2020qed} is inherently a challenging problem for term-based retrieval models. 

Our problem formulation for this part first takes inspiration from neural re-ranking \cite{xiong2017end}, where we rely on a classical yet more scalable retrieval function to identify document candidates that potentially contain the responses. The next part of the problem, response identification, closely resembles the task of question answering. Given a document $d$ and a user query $q$, we want to identify one of more text segments $\{s_i\} \in d$ such that each $s \in \{s_i\}$ can be used as a direct response to $q$. The key difference in our formulation, as we discussed in the previous section, lies within the difficulty to define a single, valid response. Previous research in argumentation community has developed a few dimensions, such as argument stance, relevance, and persuasiveness to measure the validity of a text segment being a valid argument towards a user query. Using the corpus resources developed for such tasks \cite{bar-haim-etal-2017-stance, ein2020corpus}, we can build models to score the text segments within a document along dimensions; aggregate the dimensions and retrieve a ranked list of the valid responses.

One particular challenge for some genres of web documents, such as opinion pieces \cite{liu-etal-2021-multioped}, is that the responses are usually implied instead of explicitly written. To address this, an alternative strategy is to use a text generation model to generate the responses, though it introduces additional challenge, such as evaluating the integrity of the text to the original article \cite{maynez2020faithfulness}.

\subsection{Recognizing Different Perspectives in Responses}
\label{ssec:diff_persp}
One key prerequisite for recognizing the perspectives, or semantic implications of response is to break the independence assumption of retrieved documents. When learning about controversial topic, a critical reader would naturally compare responses from various perspectives. We follow this intuition and argue for the need of a module in the document retrieval pipeline to recognize whether two responses share similar or different perspectives in the context of the query. 

Given a user query $q$ and a retrieved list of document candidates $\{d_i\}$, each with zero or more response text segments $\{s_{d_i}\}$. We want to build a similarity function $Sim(s_{d_i}, s_{d_j} | q, d_i, d_j) $. Compared to traditional sentence-level semantic similarity tasks \cite{ganitkevitch2013ppdb, cer2017semeval}, the key difference is the conditional nature of similarity here. For example, in the context of ``\emph{Should I ride bike or take taxi to work}'', ``\emph{Riding bike costs less}'' and ``\emph{Taxis cost more}'' are semantically equivalent, despite the fact that two statements have different predicate arguments. 

In practice, apart from the difficulty in developing annotations and resources for such tasks, one major bottleneck in the development of such a context-aware semantic similarity function is that text segments are typically not standalone without document as context. For example, text segments may use pronoun references instead of named or nominal mentions exclusively. To address such problems, \citet{choi2021decontextualization} proposes a text generation task to de-contextualize an input segment using its document context. 

\subsection{Trustworthiness}
\label{ssec:trust}
User trust in the results from a search engine is largely governed by the empirical \emph{authoritativeness} of their information sources and content \cite{PasternackRo13, ZhangIvRo19}. From the perspective of information sources, a source's trustworthiness is typically determined by the process it generates and publishes the content. For example, a news source is generally considered credible if all the published content go through fact-checking, and statement biases within the content  are carefully removed \cite{entman2007framing}. 

However, such information is typically neither transparent nor available from the users perspective. To increase access to information about trustworthiness of information sources, we argue for the need of two alternative strategies. First, meta information about the sources helps user recognize the type of content they are seeing, and how to assess the credibility of the content accordingly. For example, it takes drastically different approach assessing the trust value of a official guidance on COVID-19 published by a government entity, versus a opinion piece from a major news outlet criticizing the guidance. Second, for each identified or generated responses in the context of the user query, we corroborate it with supporting evidence extracted from the document \cite{aharoni2014benchmark}. These two strategies work together to provide user a potential way to visually assess the credibility of the content they see.

\begin{figure*}[t]
    \centering
    \includegraphics[width=15cm]{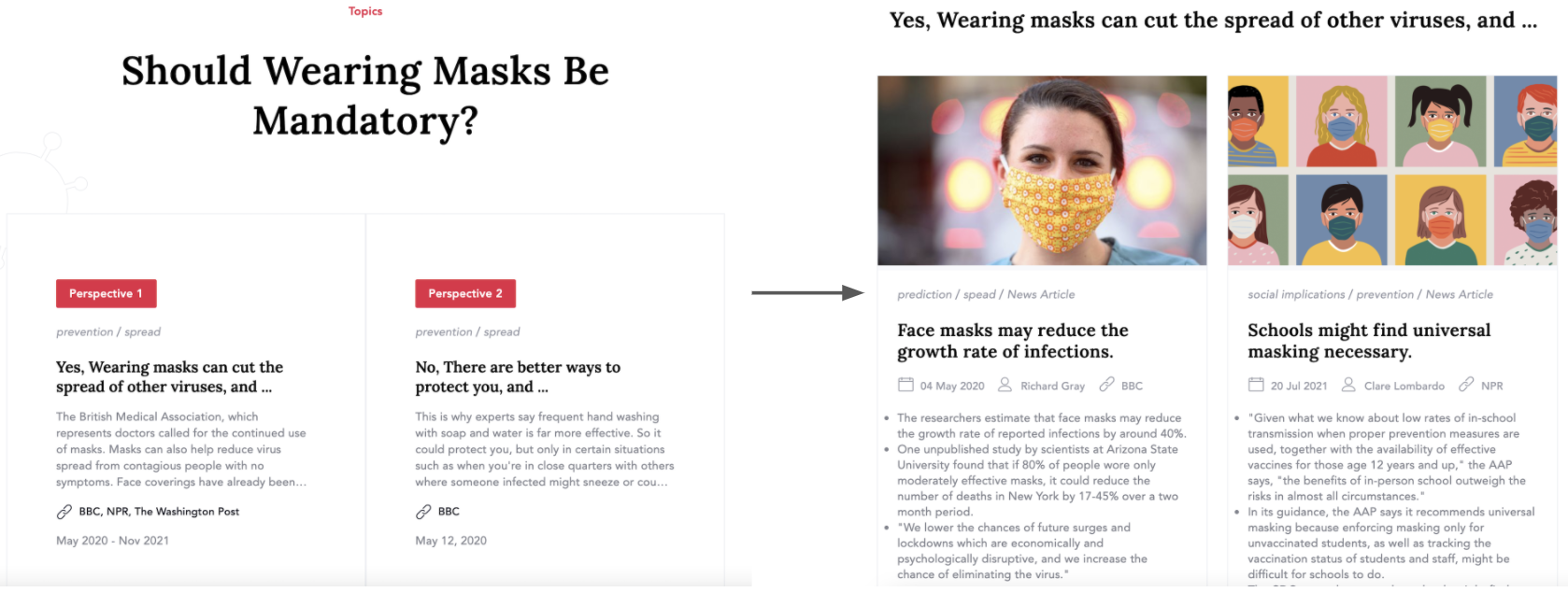}
    \caption{An example screenshot of results returned by our multi-perspective search prototype. Clicking on the Perspective 1 on the left page will direct to the page on the right. Please refer to our video (\url{https://www.youtube.com/watch?v=VD1QubNiRR4}) for a recorded live demo with the prototype search engine. See Appendix \ref{appendix:survey} for a more complete set of examples.}
    \label{fig:format}
\end{figure*}

\subsection{Putting it All Together: A Pipeline Approach}
Figure~\ref{fig:pipeline} illustrates the resulting conceptual paradigm for a multi-perspective search engine. The paradigm centers around the abstraction of a response to a query, represented by a text segment from a retrieved document.  First, a set of document candidates can be retrieved using a general document retrieval system. A response identification module ($\S$~\ref{ssec:identify_response}) extracts and validates responses from individual documents. A cross-document module ($\S$~\ref{ssec:diff_persp}) works to recognize the perspective differences between responses across documents. After corroborating evidences for each response ($\S$~\ref{ssec:trust}), the responses categorized by their perspectives, evidence from document context, along with meta information about the sources are displayed in parallel to the user. 
\label{ssec:pipeline}

\section{A Prototype of Multi-Perspective Search Engine} Following the design principles outlined in the previous section, we implement a prototype search engine. Our motivation of the prototype is two-fold. First, we want to understand the utility of our conceptual paradigm from the user perspective. Second, this prototype serves as a platform to study the user information needs with respect to controversial and debate-worthy queries. We defer the discussion of the study to $\S$~\ref{sec:user_study}.  

\subsection{Overview}

Our prototype consists of four component. First, given a user query, 
we use Google Search API to retrieve the list of top-$k$ relevant articles from the web. 
For each article, we use a query-focused summarization model \cite{liu-etal-2021-multioped} to generate the key response to the query implied by the document. 
We employ a stance classification model \cite{chen-etal-2019-seeing} to categorize the responses based on their stance (e.g. pro/con) to the query. 
Next, we use an evidence extraction model \cite{ein2020corpus} to extract sentences from each article that support the generated perspective of this article. 
Last, we cluster the articles according to their conditional semantic similarity on the query, and present them to the users with their lists of evidence. Details of each of the components are described in $\S$~\ref{sec:components}

An example search result is shown in Figure \ref{fig:format}. The user searches a query of \emph{Should wearing masks be mandatory?}, and our system presents the searched articles to the users in two clusters of perspectives, one supporting the query, and one opposing the query. In each cluster, our system displays articles that share this perspective, along with their evidence retrieved from the articles.




\begin{table}[]
    \centering
    \resizebox{1\linewidth}{!}{
    \begin{tabular}{cccc}
    \toprule
    
Task + Dataset &  System & Metric & Score\\

\cmidrule(lr){1-1} \cmidrule(lr){2-2}  \cmidrule(lr){3-3}  \cmidrule(lr){4-4}

\multirow{4}{*}{\parbox{2.5cm}{\small \centering Perspective\\ Summarization \\ \cite{liu-etal-2021-multioped}} } & \textsc{Bart} & \multirow{4}{*}{\parbox{1.5cm}{\small \centering \textsc{Rouge}$_2$}}  & 11.34  \\
& +\textsc{Rel} &  & 11.51 \\
& +\textsc{Stance} &  & 11.53 \\
& +\textsc{Rel} \& \textsc{Stance} &  & \textbf{11.92} \\
\midrule

\multirow{4}{*}{\parbox{2.5cm}{\small \centering Evidence \\ Mining \\ \cite{ein2020corpus}}} & Linear Regression & \multirow{4}{*}{\parbox{1.5cm}{\small \centering \textsc{MSE}}} & 6.39 \\
& \textsc{SVM} &  & 11.71   \\
& \textsc{GradientBoost} &  & 7.74    \\
& \textsc{BERT}$_{base}$ &  & \textbf{3.97}    \\

\midrule
\multirow{4}{*}{\parbox{2.5cm}{\small \centering Stance\\ Classification \\ \cite{chen-etal-2019-seeing}}} &  \textsc{Bert}$_{base}$ & \multirow{4}{*}{\parbox{1.5cm}{\small \centering \textsc{F1}}}  & 70.8 \\
& \textsc{Bert}$_{large}$ &  & $79.6$ \\
& \textsc{Roberta}$_{base}$ &  & $85.3$ \\
& \textsc{Roberta}$_{large}$ &  & \textbf{91.6} \\

\bottomrule 
    \end{tabular} }
    \caption{
        Evaluation results of the baselines and models on the three components. 
        All the numbers are in percentage. Top performing systems are in \textbf{bold}. The setup of different baselines in Perspective Summarization is described more in detailed in \cite{liu-etal-2021-multioped}.
    }
    \label{tab:results}
\end{table}

\subsection{Components}
\label{sec:components}
We now describe the four core components in our pipeline construction. As user interface design is not the focus of the study, we defer the discussion and more illustrative examples of our prototype website to the Appendix.
\subsubsection{Candidate Document Retrieval}
We use Google Programmable Search Engine and Custom Search JSON API\footnote{\url{https://developers.google.com/custom-search/v1/overview}} to facilitate our retrieval process. We configure our search engine to include a manually curated list of trustworthy websites to search from (See Table~\ref{tab:list_of_sources} in Appendix \ref{appendix:trustworthy} for the complete list of sources), 
and use the API to retrieve the webpages returned by our search engine. We then use Newspaper3k\footnote{\url{https://newspaper.readthedocs.io/en/latest/}} to clean and extract the article body. 

\subsubsection{Query-Focused Summarization of A Document's Perspective}
\label{sec:persp_sum}
 Given a user query and an article, we use a query-focused summarization system to extract or generate a single sentence thesis statement for the article \cite{liu-etal-2021-multioped}. 
 The generated summary is intended to serve as a \emph{direct} response to the user query. Conceptually, the generated summary should reflect the same argument that expresses the consistent stance as the article itself.
 
 We formulate the above objectives in a multi-task learning framework, as proposed by \citet{liu-etal-2021-multioped}. We start from a pretrained BART model \cite{lewis2020bart}, and finetune on the MultiOpEd dataset, an English news editorials corpus for query-focused summarization \cite{liu-etal-2021-multioped}. Following \citet{chen-etal-2019-seeing}, we use two ancillary tasks to regularize the query-focused summarization model. Specifically, we pose the \textit{relevance and stance} of the user query in the context of the article as two binary classification ancillary tasks. During the BART summarization model training, we use a relevance and a stance classification model as teacher models to score the decoded output, and train two linear classification heads on top of the BART model to mimic the predictions from the teacher models. For more details on the model training, please refer to \citet{liu-etal-2021-multioped}.
 
We conduct our evaluation in-domain on the test split of MultiOpEd dataset.  The best performing perspective summarization model achieves 11.92 Rouge-2 score on the test split of the dataset. We report the selection and performance of the baseline models in Table~\ref{tab:results}.  
 

\begin{table*}[t]
    \centering
    \resizebox{1\linewidth}{!}{
    \begin{tabular}{c|cccc}
    \toprule
    
System & Organization & Comprehensiveness & Informativeness & Overall Preference \\

\cmidrule(lr){1-1} \cmidrule(lr){2-2}  \cmidrule(lr){3-3}  \cmidrule(lr){4-4} 
\cmidrule(lr){5-5}

Google & 92 (31\%) & 78 (26\%) & 135 (45\%) & 110 (37\%) \\
Multi-Perspective Search & \textbf{208 (69\%)} & \textbf{222 (74\%)} & \textbf{165 (55\%)} & \textbf{190 (63\%)} \\

\bottomrule 
    \end{tabular} 
    }
    \caption{
        Number of total votes 
        from 10 participants comparing Google's search results to our prototype's on 30 different controversial queries. For each query, we ask participants to vote for the preferred system (i.e. Google vs. ours) from four criteria: (i) Organization of search results, (ii) Comprehensiveness of search results as potential answers to the query, (iii) Informativeness of search results, and (iv) Overall preference . See Appendix \ref{appendix:hypo_test} for statistical significance testing and more details on the result.
    }
    \label{tab:survey}
\end{table*}

\subsubsection{Query-Conditioned Similarity Between Perspectives}
Next, our system is expected to group articles with similar perspectives within the context of the query. Following \citet{chen-etal-2019-seeing}, we conceptually treat the problem as conditional similarity between two perspectives. Given a query, and two single-sentence perspective summaries, we learn a conditional similarity function for whether the two perspectives present similar arguments in the context of the query. 

For our experimental setup, we finetune a pretrained RoBERTa-base model plus a linear layer on the stance classification sub-task in the Perspectrum dataset. We follow the same experimental setup in \citet{chen-etal-2019-seeing}.  Our stance classifier achieves 91.6\% accuracy on the test split of the Perspectrum dataset. 
Given a user query and a generated perspective as input to the stance classification model, we use the last hidden layer output from the backbone RoBERTa model as the representation of the perspective conditioned on the user query. For all perspectives generated, with their corresponding articles for a user query, we use k-means clustering with $k=3$ on the encoded representations. Conceptually, as we use a stance classification model as our encoder, the clusters resemble the different stances that responses can take towards the user query.

\subsubsection{Identifying Evidence from Document}
To corroborate the argument presented in the generated perspective, we extract up to three sentences from the article as supporting evidence. We formulate this as a ranking task. Given a generated perspective and all sentences from the corresponding document, we train a regression model to predict the probability of each sentence being evidence towards the perspective. We choose Evidence Sentences dataset \cite{ein2020corpus}, an English argument mining dataset that most closely resembles our task. We finetune a pretrained BERT-base model with 768 hidden size plus a linear layer to predict the probability of a sentence being evidence towards the perspective. The training takes less than 30 minutes on a GeForce 1080Ti GPU and we conduct our experiments in a train/dev/test setting of 80\%/10\%/10\%. 

For each document and its perspective, we extract a ranked list of sentences with top-$k$ highest output probabilities from the model. As the training objective and data does not pose any constraint on whether the evidence supports or opposes the input perspective, we pipe it with a stance classifier \cite{chen-etal-2019-seeing} to select only the sentences that support the perspective, and use a relevance classifier \cite{chen-etal-2019-seeing} to rank them and present the three most relevant evidence paragraphs to the user. 
We conduct our evaluation on the test split of Evidence Sentences dataset, and our evidence mining system achieved a mean squared error of $3.97\%$.



\section{A Study on User Information Needs}
With the prototype search engine, we revisit the discussions in $\S$~\ref{sec:design_challenge}. Our goal for the user study is to validate our hypothesis on user information need, and discover new insights from users' experience with an actual prototype model. 
\label{sec:user_study}
\subsection{Setup}
We collect and construct 30 queries of 6 controversial topics from 2020 United States presidential debate\footnote{https://2020election.procon.org/}. The 6 topics include  COVID-19, Race \& Justice, Immigration, Healthcare, Climate change, and Policy. We conduct a survey that compares the search results of these queries returned by Google search with the results returned by our prototype. The complete list of queries can be found in Appendix~\ref{appendix:queries}.

We use a Google Programmable Search Engine to retrieve the same list of ten articles for each query. Our prototype system would further process the articles and present users with extracted responses, evidence paragraph in different order, i.e. organized based on the predicted stance. For each controversial query, we show ten participants five different questions that compare the retrieved results from Google with those from our prototype. For each query and retrieved results from two systems, we ask the participants to choose which system (i) provides a better organization of results; (ii) shows a more comprehensive view of the topic; and (iii) is more informative in the context of the topic. We then ask for their the overall preferences between the system, and their explanations, expressed in free form text, for their choices. 

To eliminate user interface (UI) as a confounding factor for the study, we construct a web interface for the survey that presents two sets of results with nearly identical UI. A screenshot of the survey interface is shown in Appendix \ref{appendix:survey}. We hire participants from Amazon Mechanical Turk to collect responses. Participants are located in the United States and are required to have master qualifications, i.e. top performers recognized by MTurk among all workers. We compensate \$1 for answering 25 questions in total for five queries. The compensation rates are determined by estimating the average completion time for the survey. The participants are anonymized using the Amazon Mechanical Turk ID. The data collection protocol is determined exempt by an ethics review board.

\subsection{Quantitative Analysis}

The statistics from our survey responses are shown in Table \ref{tab:survey}. $63\%$ of the responses indicate that they generally prefer our prototype to Google Search on the query. $69\%$ and $74\%$ of the responses state that our prototype has a better organization of the results and offers a more comprehensive view of the query. The noticeable advantages of these two aspects align with our assumptions. We believe that users prefer seeing search results in different clusters of perspectives instead of in a list ranked by relevance, and are convinced that such organization presents users a broader and more diverse set of views of the topic. On the other hand, only $55\%$ of the responses indicate that our prototype presents more information on the topic. This meets our expectation as well given that the two systems are presenting the same set of articles for each query.

We perform hypothesis testing to validate that users prefer our prototype over Google Search on all of the aspects. We resample from the distribution of the results of each question by boostrapping and compute the mean and standard deviation of each distribution. We repeat this process for 1000 times and use the average mean and standard deviation as the sample mean and sample standard deviation for each question. We then perform one-tailed z-test for each of the sample distribution. The p-values of all aspects except Informativeness are smaller $0.01$, giving us $99\%$ confidence to reject the null hypothesis that less than half of the participants prefer our prototype under on these aspects, and the p-value of Informativeness is $0.043$, showing a $95\%$ confidence level in improvements over Google Search.

\subsection{Qualitative Analysis}
To understand the information needs of participants, we look into the explanations they provided for their general preferences on the system. We find that some participants prefer the search results provided by our prototype because they seek for direct answers to the query: \emph{The search is asking a question, and the results of System 1 (Google) offer only a collection of articles without any real answers. I like that System 2 (Our prototype) filters the results by organizing the "yes" articles together, the "no" articles together, and even the "maybe" articles together. This is a more convenient and helpful way for the user to find what they're really looking for.} This finding confirms our hypothesis that some users expect to see direct answers when searching for a controversial query. 

Some other participants prefer the way we present the information by showing bullet points instead of snippets: \emph{I prefer system 2 because each result has bullets of information which makes it easier for me to decide if it is worth clicking. Also, it is separated by results that answer yes/no to the question.}. It shows that some users prefer to see an organized list of key information in an article and use it to decide whether or not the article is worthwhile reading. 

Despite our intent of providing a diverse set of perspectives to alleviate selection bias, a small portion of participants express concern that the automation provided by our prototype would in turn create biases comparing to Google's. \emph{I prefer system 1 because it gives a more balanced set of articles and gives the impression of being more unbiased, whereas System 2 has a statement right at the beginning that lends one to think of bias.} Even when given the exactly same set of articles, a participant thinks that our prototype gives an impression of being more biased because our system explicitly states the perspective of an article, whereas Google Search results don't show that explicitly and require the users to look into the articles to find out. This finding shows that the information needs of users may vary, and explicitly displaying the stances of articles may have biased connotations. 

\section{Discussion}
The results from the user study offer validation our main hypothesis for the paper, that users typically wish to see \emph{direct and diverse} responses when issuing a controversial, debate-worthy query to the search engine. At the same time, we also see concerns over potential model mistakes and intrinsic biases, which lead to decreased trust compared to existing document retrieval services. 

Surrounding the three design factors introduced in $\S$~\ref{sec:design_challenge}, we discuss a few existing trade-offs in our conceptual paradigm through the lens of technical challenges involved in developing a multi-perspective search system.   

\paragraph{Direct Response vs. Trust}
As the responses act as an extra layer of abstraction over the documents, the question on the trustworthiness of the machine-identified response naturally rises as a concern. 

Such concerns involve two conflicting factors of research and design considerations. First, explanation-driven design of natural language understanding tools is beneficial, not only for improvements on model performance, but allows for more efficient design iteration of models. For this reason, explanation-driven task and data design has drawn more attention for information retrieval applications \cite{lamm2020qed}.

However, from the user experience point of view, the harm of displaying the wrong explanation might overwhelm the benefit of increased model performance by incorporating explanation. In the case of controversial queries, users are expected to be more sensitive to such trade-offs. 

\paragraph{Diverse Response vs. Trust} An unfortunate yet existing factor that influences user trust is \emph{confirmation bias}. For example, \citet{meppelink2019right} discovers in the domain of health information retrieval, the user tend to only agree with results that conform with their prior belief. This raises a question about whether displaying a diverse set of perspectives can influence or correct the potential biases within users' prior belief.

Assessing such hypothesis is out of scope of this current project, and we will defer a study of such to the future. However, as empirical evidence show that such confirmation bias can be attributed in part to the users' access pattern and consumption of news information \cite{knobloch2015confirmation}, we tend to believe that improving how search engines present controversial information will have a positive impact on alleviating confirmation bias among consumers of news information. 

\label{sec:discussion}

\section{Conclusion}
This study aims to understand the user information needs for controversial and debate-worthy queries to search engines. We argue for and examine the need of delivering \emph{direct and diverse} response to such queries. To demonstrate the utility of the paradigm, we develop a prototype multi-perspective search engine by synthesizing a suite of relevant NLP and IR tasks. We use the tool to facilitate a user study that confirms the benefit of  serving \emph{direct and diverse} responses to user queries. Through the positive and negative feedback we receive on the prototype, we discuss a few trade-off between technical design and user experience. We hope the findings in the paper offers insights, guidance and opportunities for future development on web document retrieval systems.

\section*{Ethical Considerations}
As the data and models used to develop the multi-perspective search engine inevitably contain artifacts and inaccuracy,  the returned results should only serve as speculation, as opposed to consolidated evidence for user queries, and should not be taken as granted without further reasoning and validation. The participants of the user study are informed of the potential risks of such before the annotation task. 

\section*{Acknowledgments}
The authors would like to thank Disha Jindal, Hegler Tissot for helpful conversations throughout the development of the prototype search engine, and the anonymous reviewers for their valuable feedback on the draft. In addition, the authors are grateful for the collaboration with \texttt{10clouds} \footnote{\url{https://10clouds.com/}} to research, design and develop the front-end interface for the prototype. This work was supported in part by a Focused Award from Google, and in part by the Oﬃce of the Director of National Intelligence (ODNI), Intelligence Advanced Research Projects Activity (IARPA), via IARPA Contract No. 2019-19051600006 under the BETTER Program. The views and conclusions contained herein are those of the authors and should not be interpreted as necessarily representing the oﬃcial policies, either expressed or implied, of ODNI, IARPA, the Department of Defense, or the U.S. Government. The U.S. Government is authorized to reproduce and distribute reprints for governmental purposes notwithstanding any copyright annotation therein.





\bibliography{anthology,custom, ccg}
\bibliographystyle{acl_natbib}

\newpage
\appendix

\section{Hypothesis Testing}
\label{appendix:hypo_test}

Our null hypothesis is $P\leq P_0$ and alternate hypothesis is $P > P_0$, where $P_0=0.5$ and $P$ is the percentage of responses of the survey that prefer our prototype.

We use the one-sided Z-Test as our hypothesis testing statistics. The z-score of a standard z-test can be calculated as $$Z = \frac{\bar{x}-\mu}{\frac{\sigma}{\sqrt{N}}}$$

We perform Boostrapping resampling to get the sample distribution and standard deviation. Specifically, we randomly sample data points with replacement from the distribution of each question's results to a sample size of 300, and repeat this process $1000$ times to compute an average mean and average standard deviation. We then regard these mean and standard deviation as the sample mean and sample standard deviation. We then use these statistics to calculate the z-score for each type of question and get  $z = 7.29$, $9.57$, $1.72$, and $4.79$. We show that users prefer our prototype over Google Search with significance levels 0.01, 0.01, 0.05, and 0.01 for questions on \emph{organization}, \textit{comprehensiveness}, \textit{informativeness}, and \textit{general preference}, respectively.


\section{List of Sources}
\label{appendix:trustworthy}
See Table \ref{tab:list_of_sources}.
\begin{table}[h]
    \centering
    \begin{tabular}{|c|}
         \hline
         nytimes.com\\
         theguardian.com \\
         npr.org \\ 
         reuters.com \\
         bbc.com \\
         cnn.com \\
         washingtonpost.com \\
         nbcnews.com \\ 
         cbs.com \\
         abc.com \\
         foxnews.com \\
         who.int \\
         clevelandclinic.com \\
         medicalnewstoday.com \\
         noah-health.org \\ 
         familydoctor.org \\
         medlineplus.gov \\
         mayoclinic.org \\
         webmd.com \\
         cebm.net \\
         sciencenews.org \\
         sciencemag.org \\
         yalemedicine.org \\
         nejm.org \\
         \hline
    \end{tabular}
    \caption{List of urls of the sources we include as information source for our prototype multi-perspective search engine. }
    \label{tab:list_of_sources}
\end{table}

\section{List of Controversial Queries}
\label{appendix:queries}
See next pages.
\begin{table*}[]
    \centering
    \begin{tabular}{|c|c|}
        \hline
         Topic & Query \\
         \hline
        \multirow{5}{*}{\parbox{3cm}{{\centering COVID-19}}} & Should wearing masks be mandatory? \\ 
        &  Should we all get vaccinated? \\
        &  Is herd immunity for COVID-19 achievable? \\
        &  Should COVID-19 vaccines be mandatory? \\
        &  Will the Covid-19 pandemic have a lasting impact on society? \\
        \hline
        \multirow{5}{*}{\parbox{3cm}{{\centering Race \& Justice}}} & Sshould the US ban assault weapons? \\ 
        & Should police departments be defunded, if not abolished? \\
        & Should the death penalty be allowed? \\
        & Should the use of private prisons continue? \\
        & Should the US continue to build a border wall at the US-Mexico border? \\
        \hline
        \multirow{5}{*}{\parbox{3cm}{{\centering Immigration}}} & Will immigrants take american jobs? \\ 
        & Should the US end the Deferred Action for Childhood Arrivals (DACA) policy? \\
        & Should the US decriminalize illegal border crossings? \\
        & Should immigrants who entered the US illegally be denied a path to citizenship?\\
        & Should America use a merit-based immigration system? \\
        \hline
        \multirow{5}{*}{\parbox{3cm}{{\centering Health Care}}} & Should we support medicare for all? \\ 
        & Does racial inequality persists in health care coverage? \\
        & Should abortion be legal? \\
        & Do the prescription drug costs need to be lowered?\\
        & Should imported prescription drugs be allowed in the U.S.? \\
        \hline
        \multirow{5}{*}{\parbox{3cm}{{\centering Climate Change}}} & Should the US rejoin the Paris climate agreement? \\ 
        & Should the US adopt a climate change plan? \\
        & Should the US expand fossil fuel extraction on public land? \\
        & Should fighting climate change be a priority?\\
        & Is climate crisis inevitable? \\
        \hline
        \multirow{5}{*}{\parbox{3cm}{{\centering Policy}}} & Should the US have withdrawn from the Open Skies Treaty? \\ 
        & Should the tariffs imposed on China by president Trump be maintained? \\
        & Should confederate statues be taken down? \\
        & Should the federal government adopt Net Neutrality Rules?\\
        & Should the US re-enter a nuclear deal with Iran? \\
        \hline
        
    \end{tabular}
    \caption{List of controversial topic and queries we include in our user study. The topics and queries are selected from the 2020 United States presidential debates. The list of topics are selected from \url{https://2020election.procon.org/}. }
    \label{tab:my_label}
\end{table*}

\section{Screenshots of Demo Website and Human Study Interface}
\label{appendix:survey}

\begin{figure*}[t]
    \centering
    \includegraphics[width=15cm]{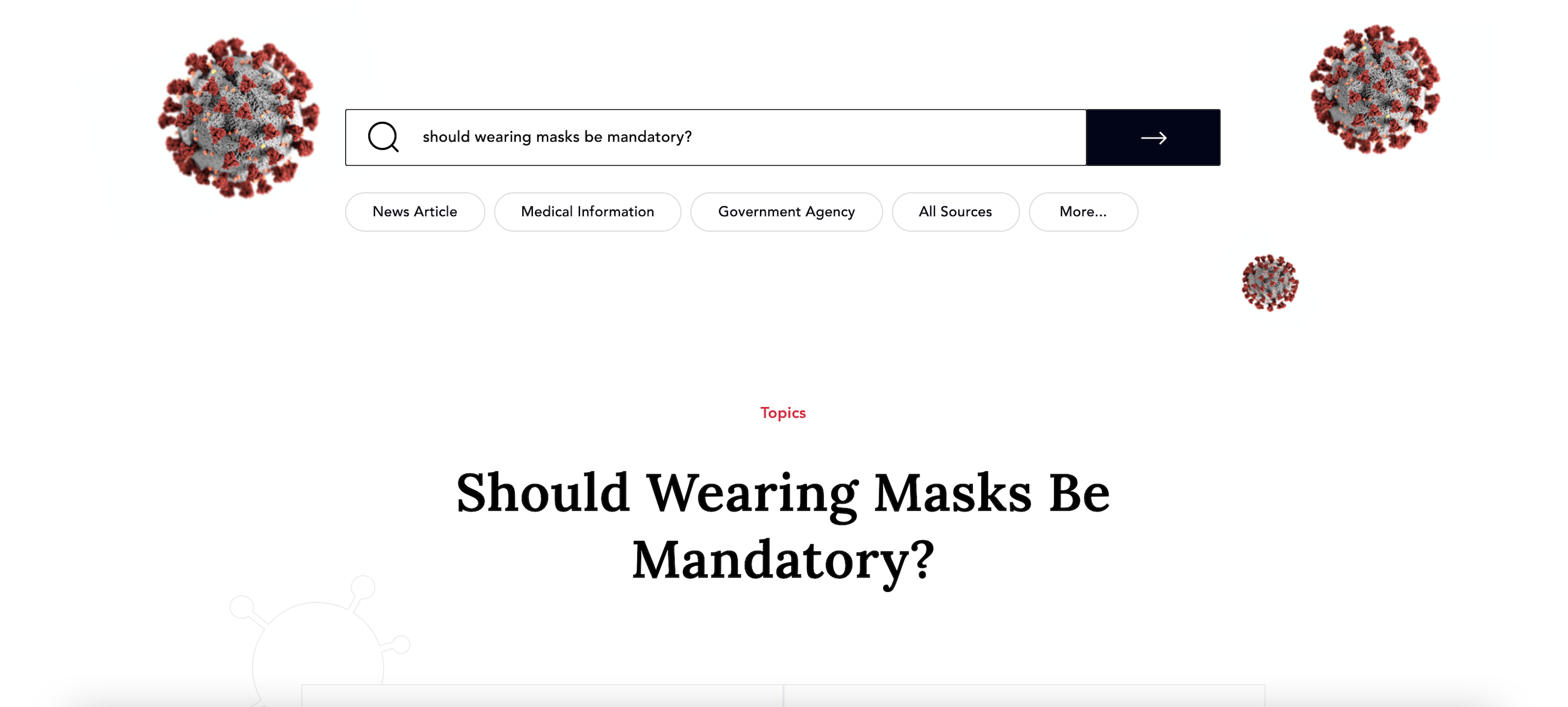}
    \caption{Screenshot 1 of our prototype's demo website.}
    \label{fig:survey}
    \centering
    \includegraphics[width=15cm]{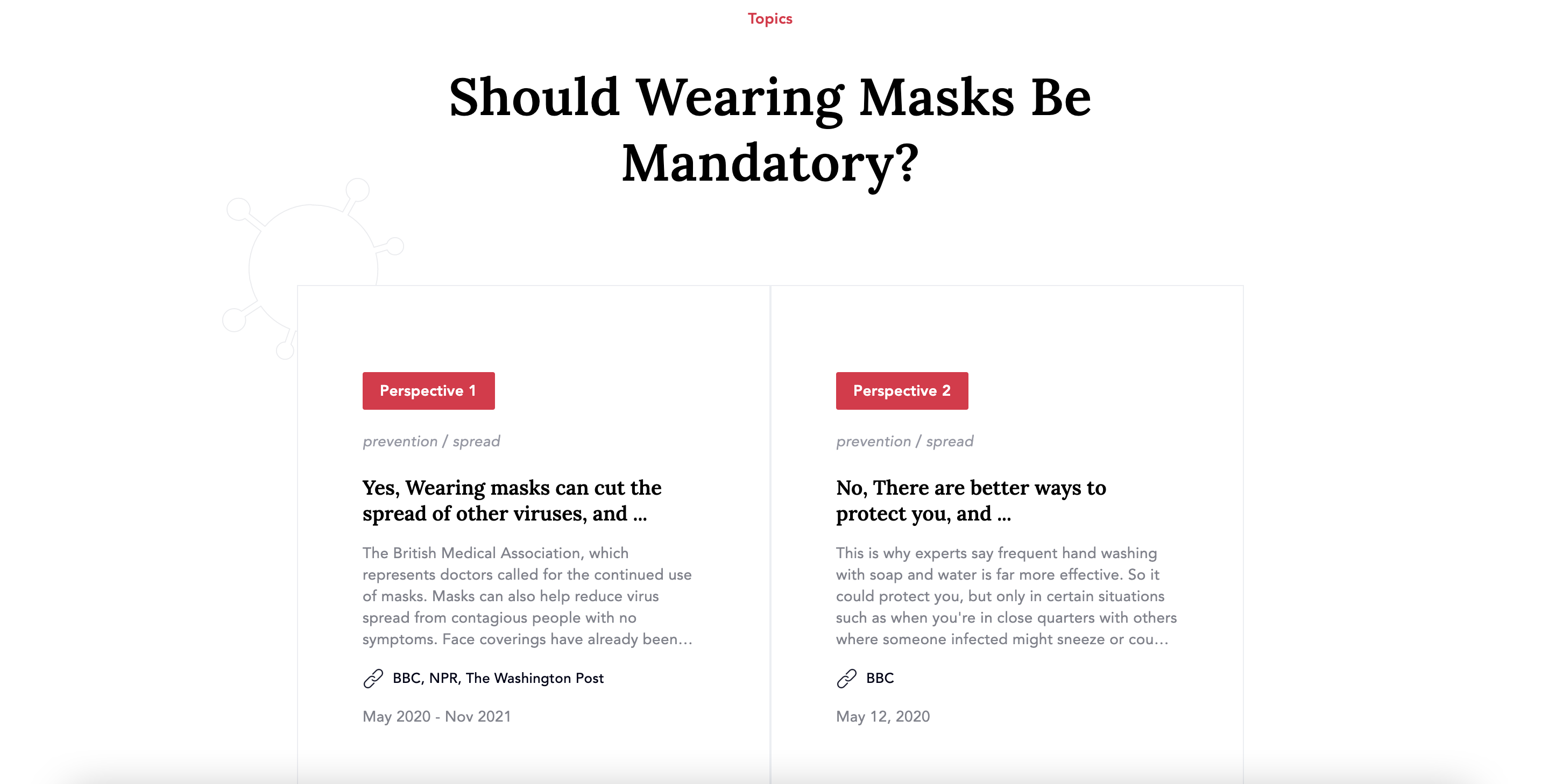}
    \caption{Screenshot 2 of our prototype's demo website.}
    \label{fig:survey}
    \centering
    \includegraphics[width=15cm]{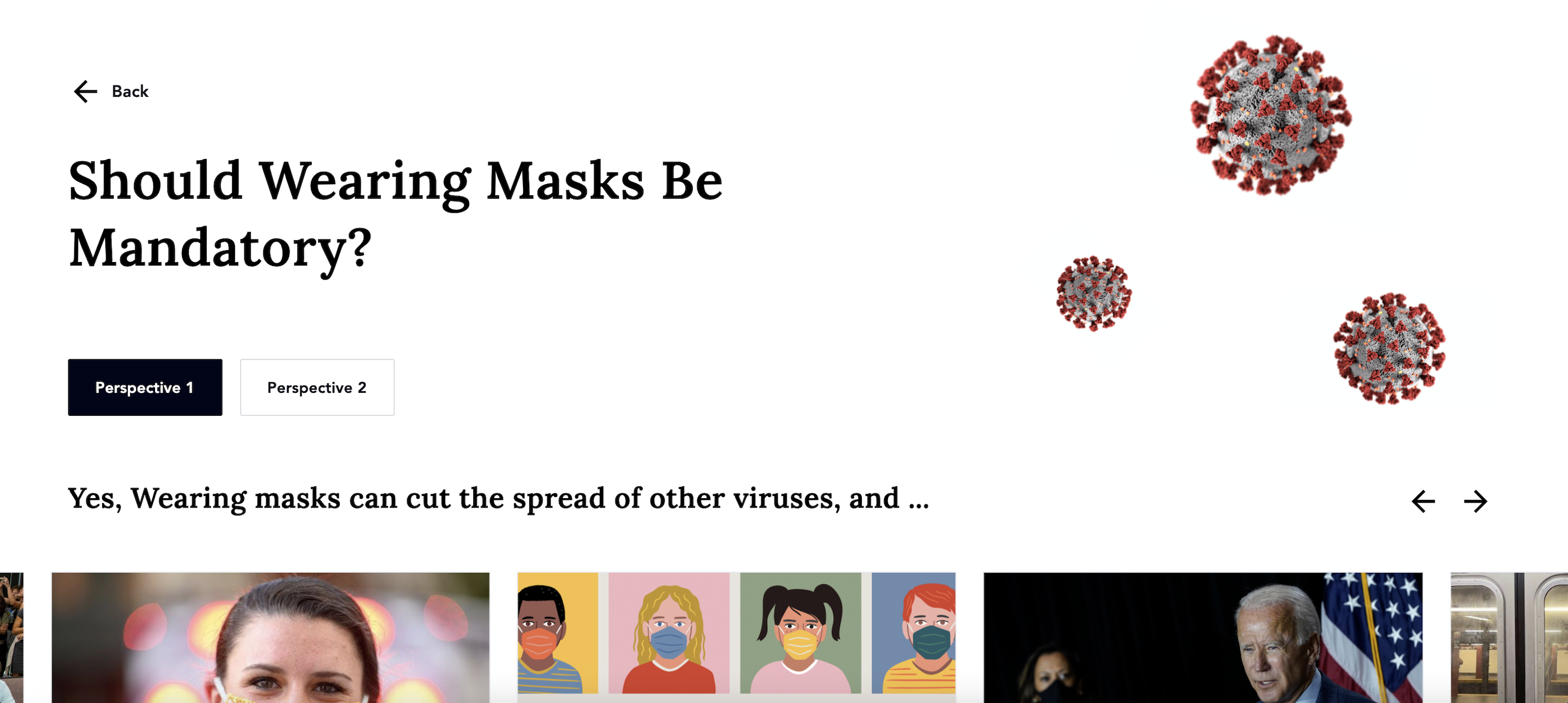}
    \caption{Screenshot 3 of our prototype's demo website.}
    \label{fig:survey}
\end{figure*}

\begin{figure*}[t]
    \centering
    \includegraphics[width=15cm]{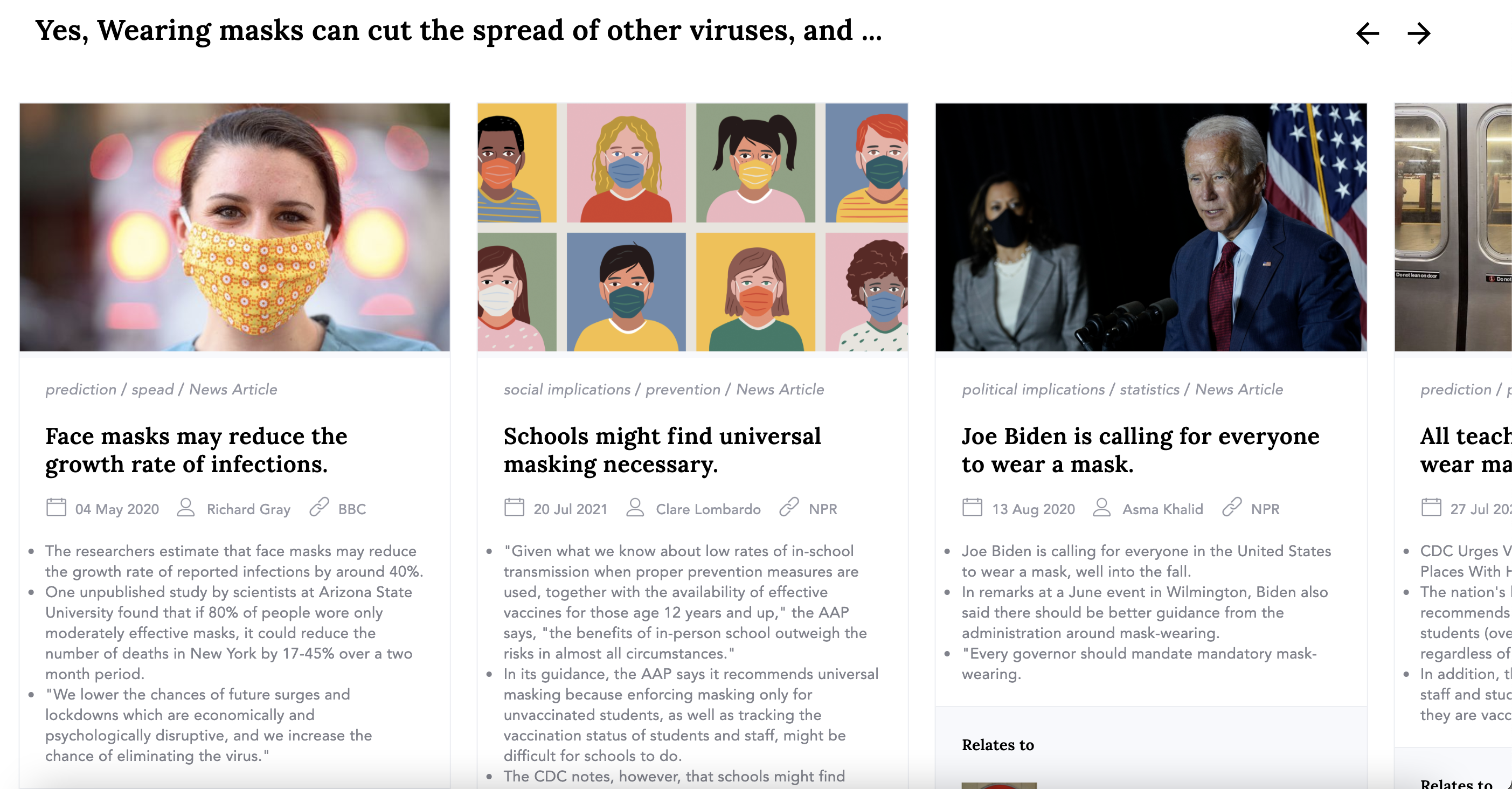}
    \caption{Screenshot 4 of our prototype's demo website.}
    \label{fig:survey}
    \includegraphics[width=15cm]{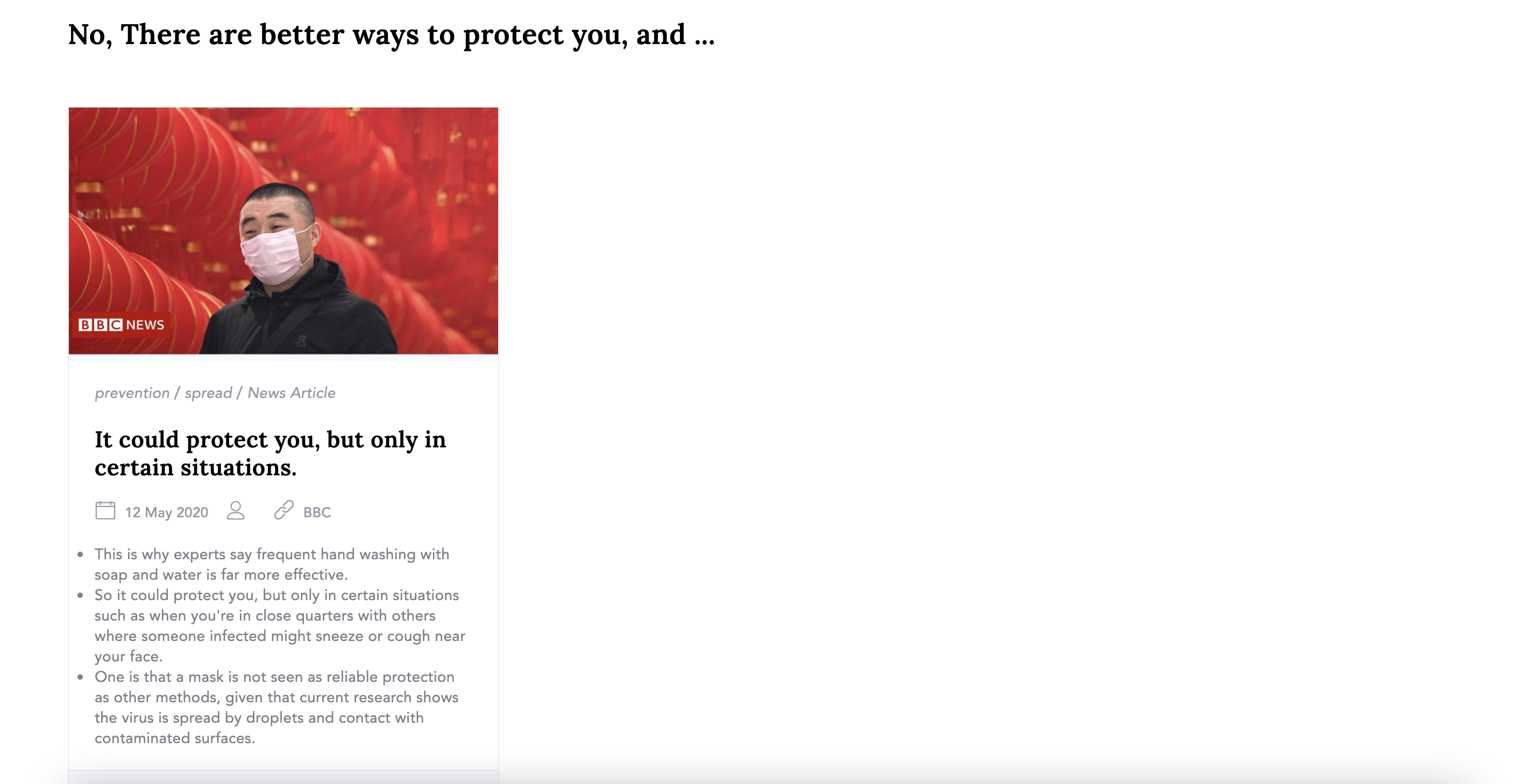}
    \caption{Screenshot 5 of our prototype's demo website.}
    \label{fig:survey}
\end{figure*}

\begin{figure*}[t]
\centering
    \includegraphics[width=15cm]{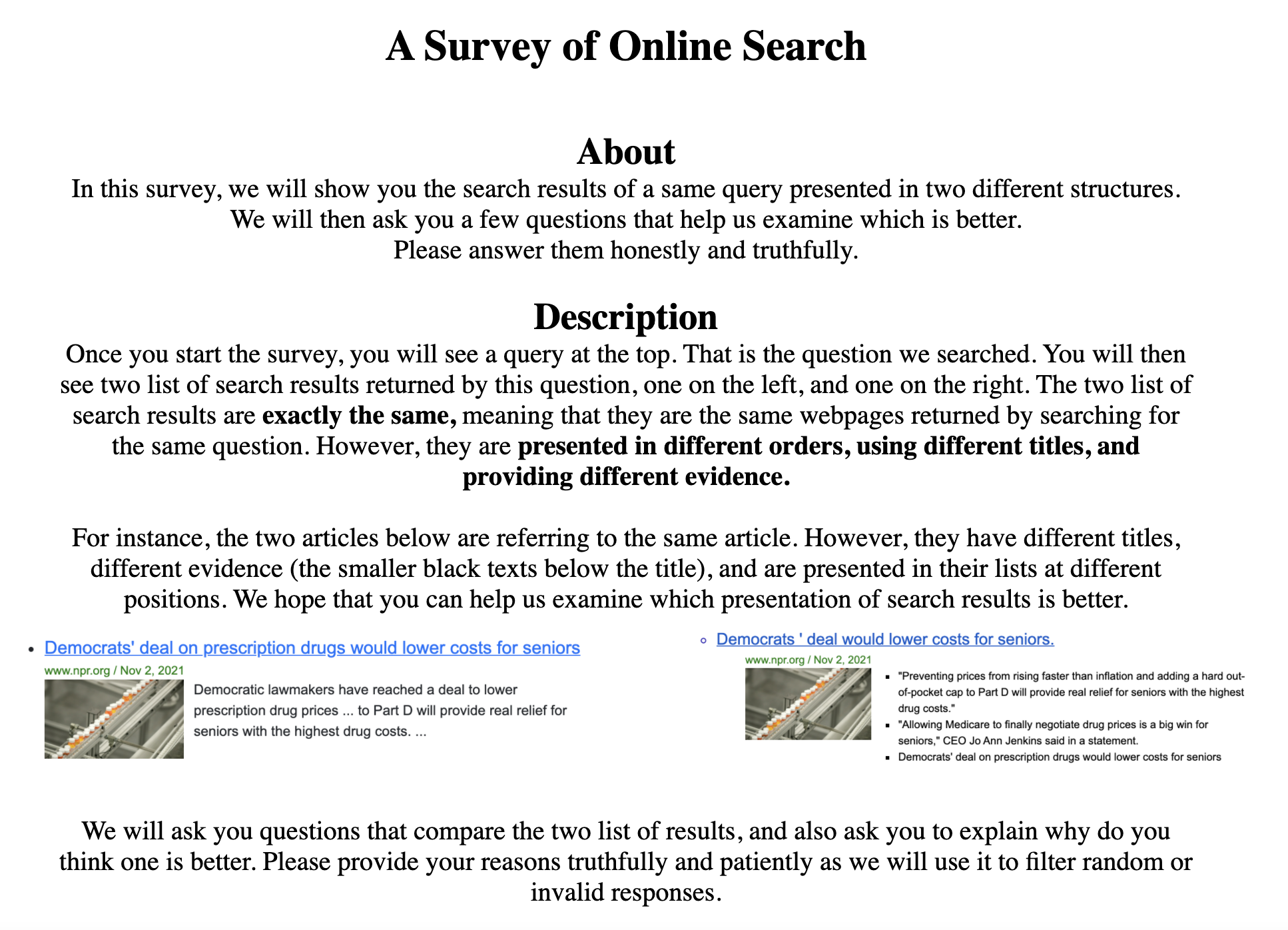}
    \label{fig:survey}
    \centering
    \includegraphics[width=15cm]{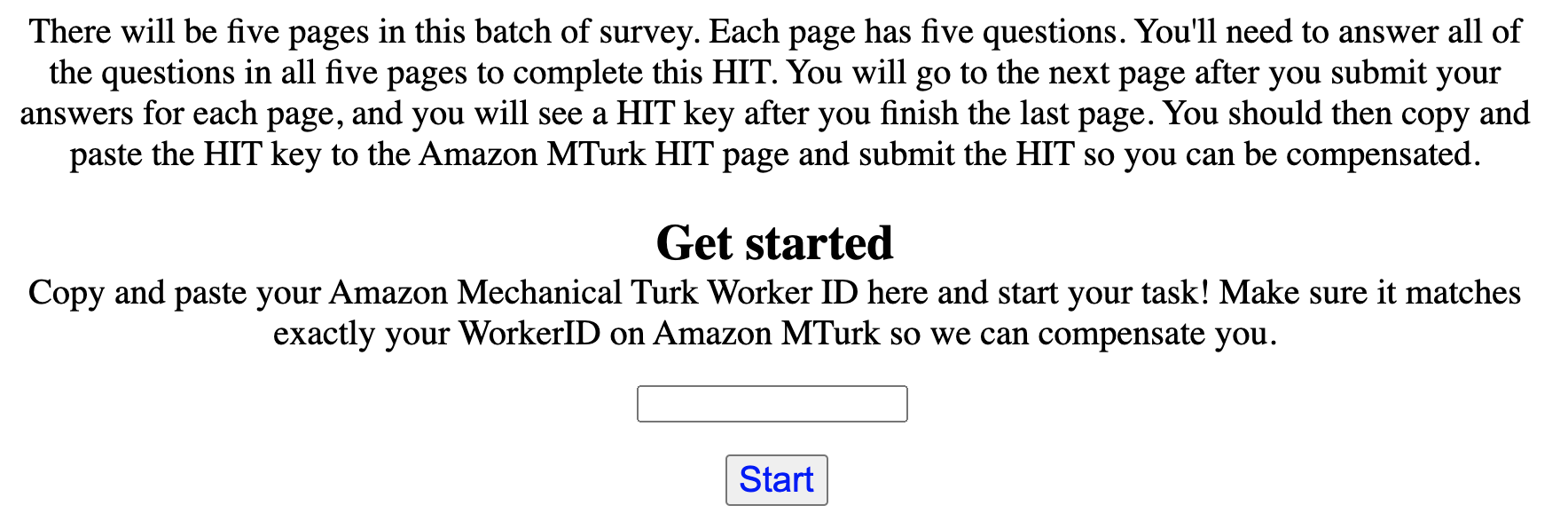}
    \caption{Screenshot 1 of the survey interface.}
    \label{fig:survey}
\end{figure*}

\begin{figure*}[t]
    \centering
    \includegraphics[width=12.5cm]{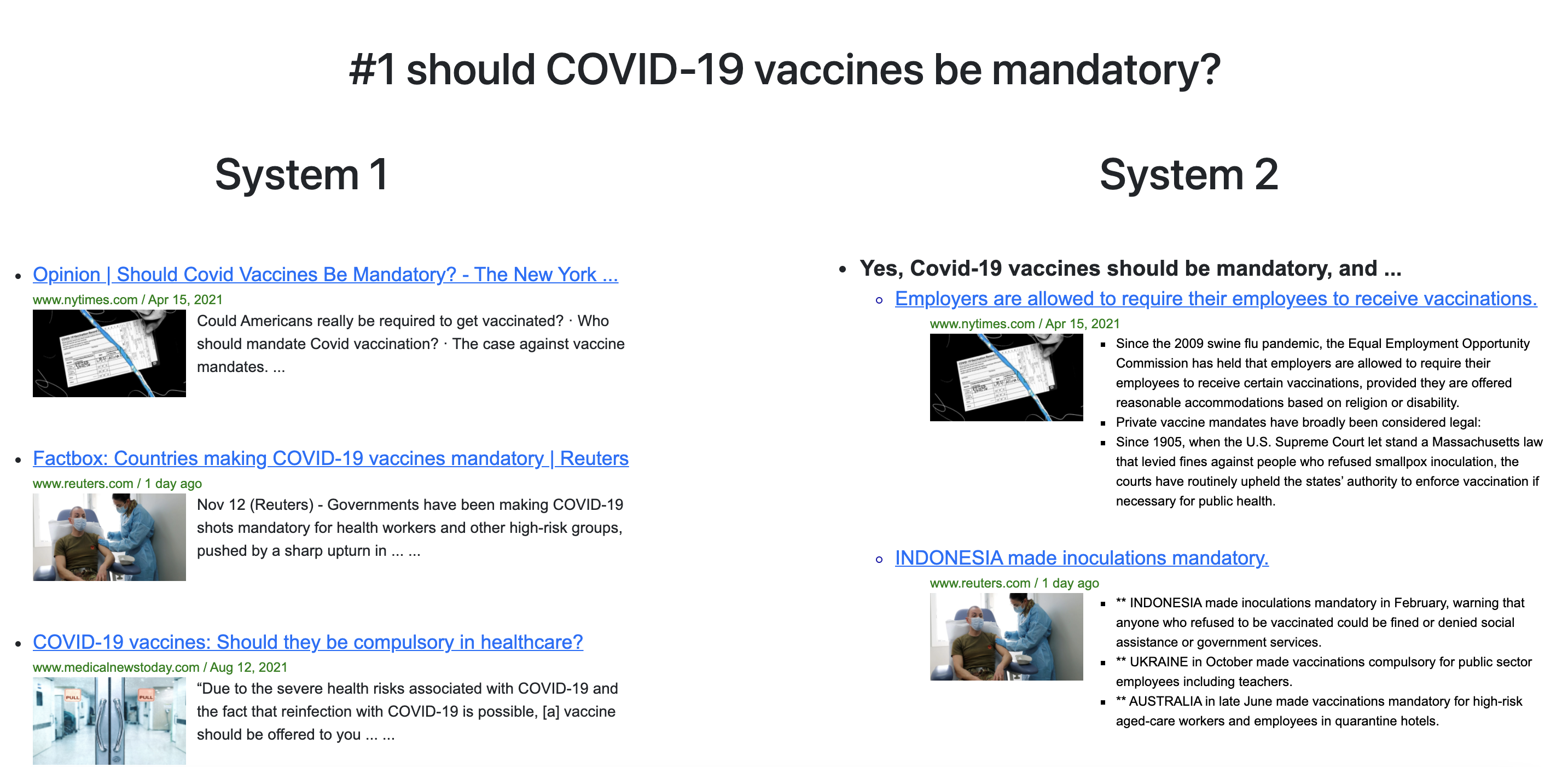}
    \label{fig:survey}

    \centering
    \includegraphics[width=12.5cm]{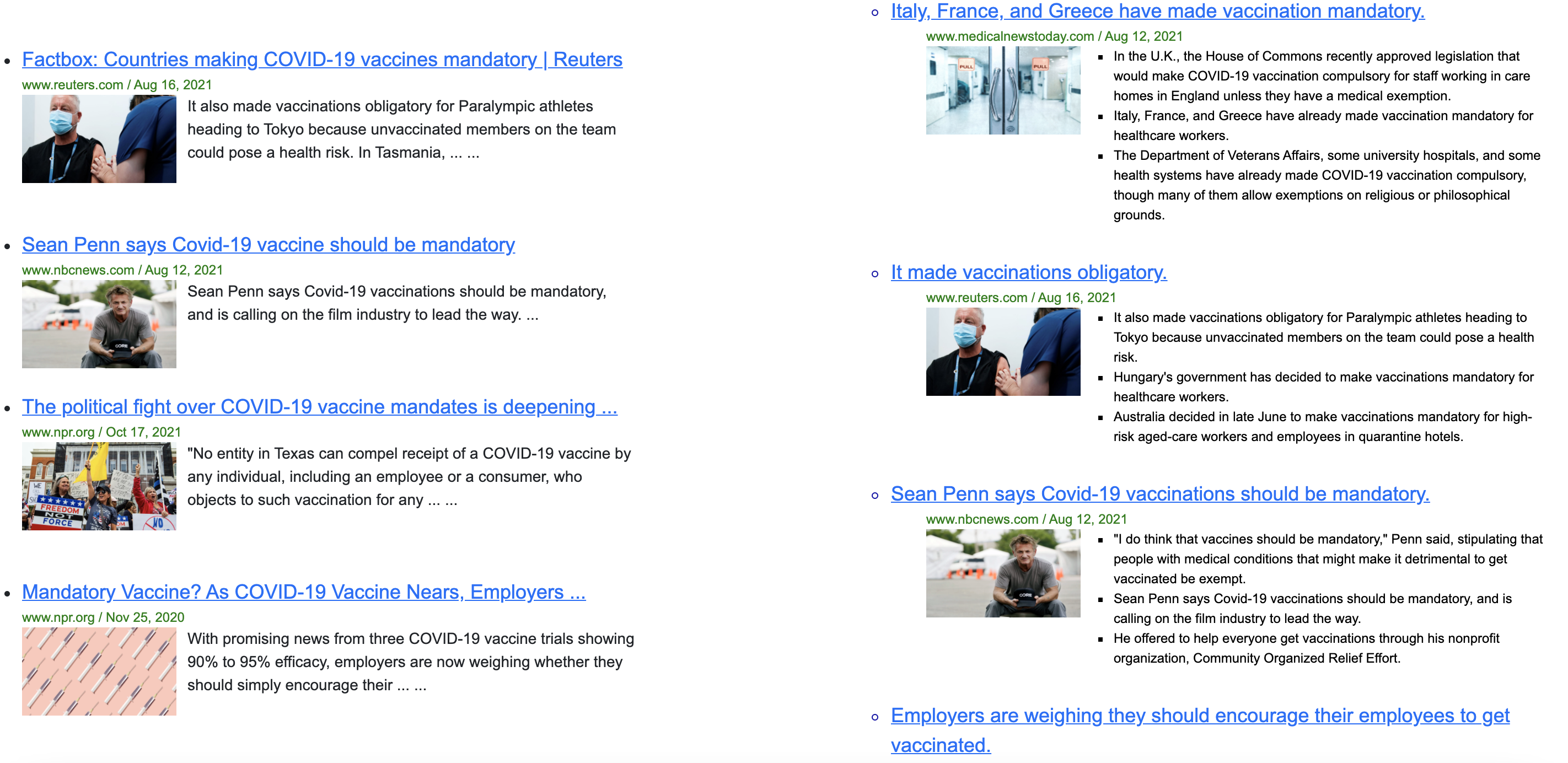}
    \label{fig:survey}
    
    \centering
    \includegraphics[width=12.5cm]{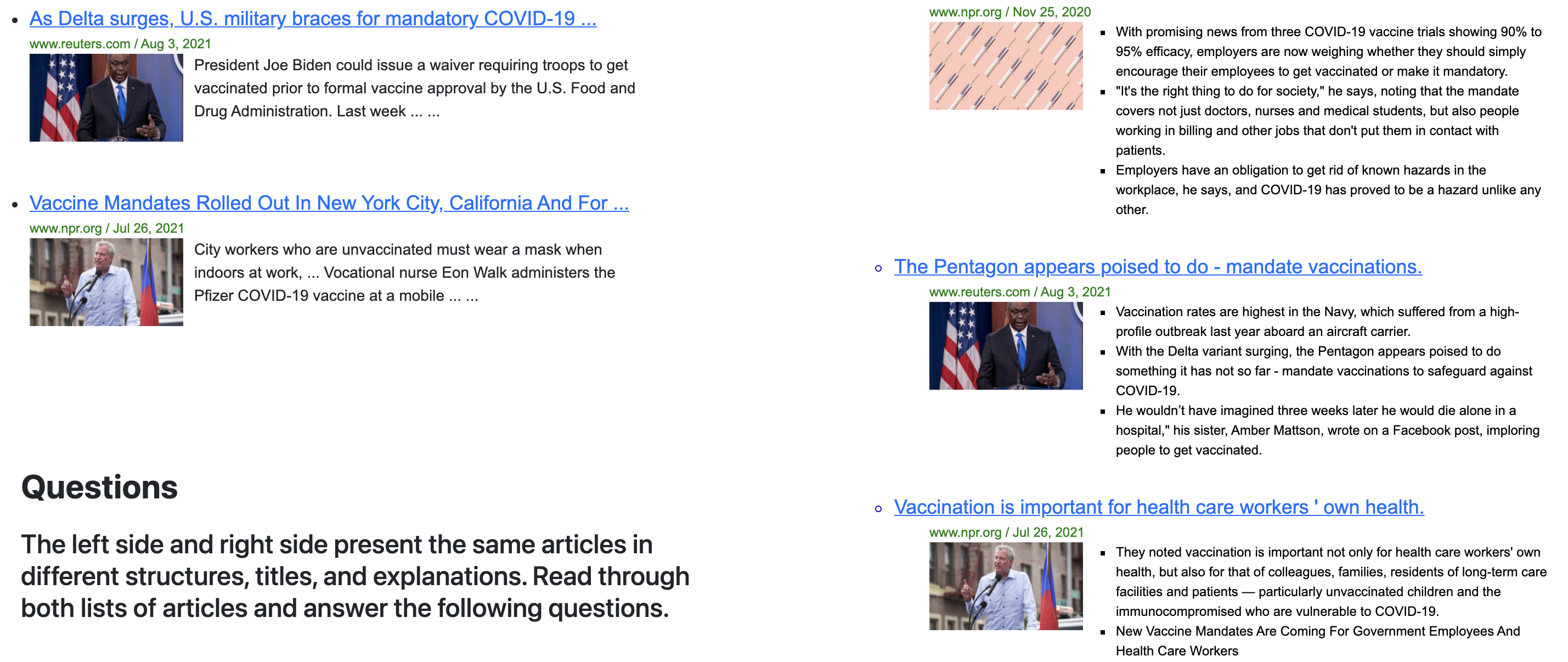}
    \label{fig:survey}
    
    \centering
    \includegraphics[width=12.5cm]{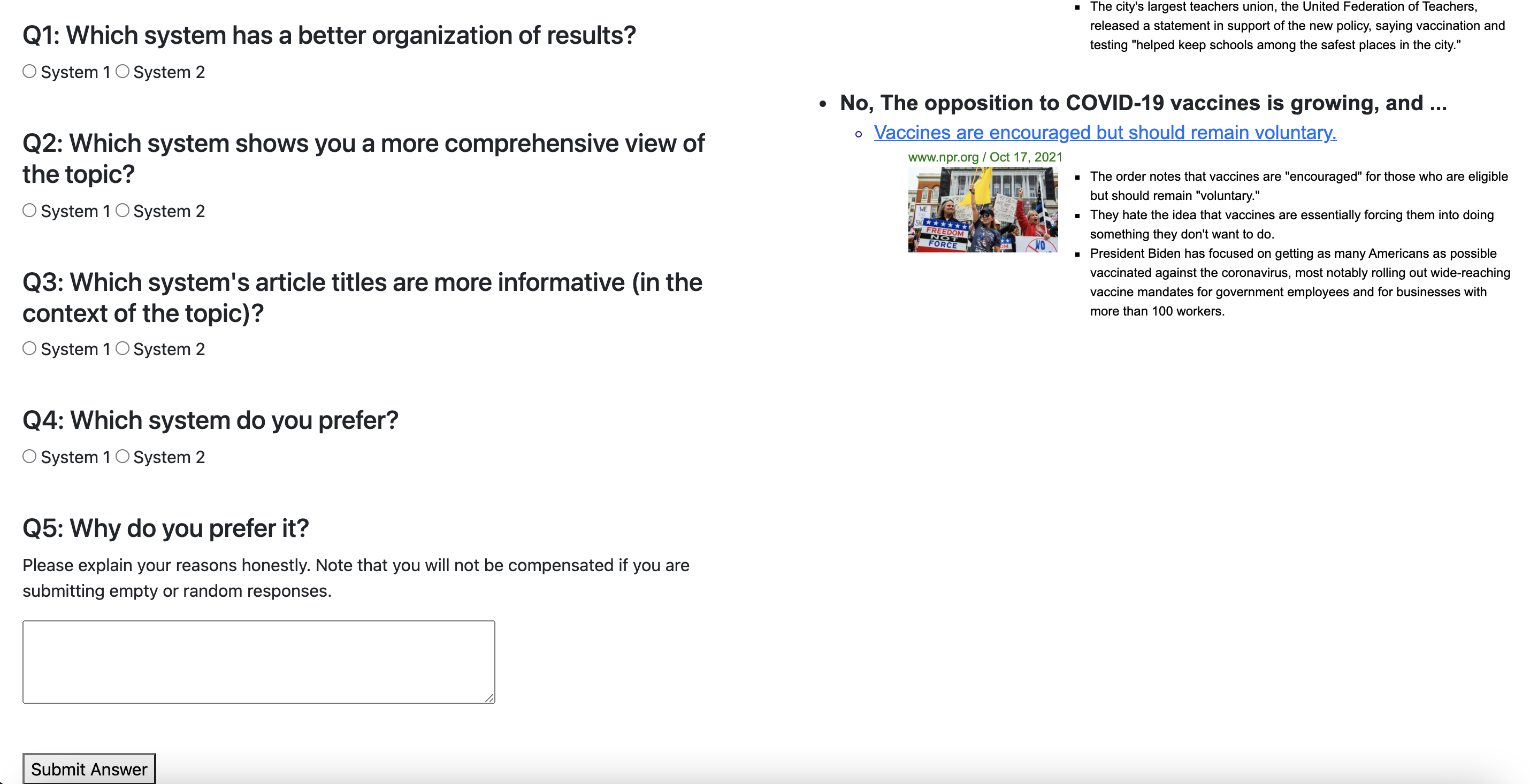}
    \caption{Screenshot 2 of the survey interface.}
    \label{fig:survey}
\end{figure*}

\end{document}